\documentclass[runningheads]{llncs}
\usepackage[T1]{fontenc}
\usepackage{graphicx}
\usepackage{multicol}
\usepackage[hidelinks]{hyperref}
\usepackage{multirow}
\usepackage{amsmath}
\usepackage{amssymb}
\usepackage{amsfonts}
\usepackage{cite}
\usepackage{xcolor}
\usepackage{booktabs}
\usepackage{algorithm}
\usepackage[noend]{algpseudocode}

\begin{document}
\title{Reliable Semantic Understanding for Real World Zero-shot Object Goal Navigation}
\titlerunning{Reliable Semantic Understanding for Real World ZS-OGN}
\author{Halil Utku Unlu*\inst{1}\orcidID{0000-0002-4368-5172} \\\and
Shuaihang Yuan*\inst{2,3,4}\orcidID{0000-0002-7092-7966}\thanks{* denotes the equal contribution}\\ \and
Congcong Wen\inst{2,4}\orcidID{0000-0001-6448-003X} \and
Hao Huang\inst{2,4}\orcidID{0000-0002-9131-5854} \and
Anthony Tzes\inst{2,3}\orcidID{0000-0003-3709-2810} \and
Yi Fang\inst{2,3,4}\orcidID{0000-0001-9427-3883}
}
\authorrunning{H. U. Unlu et al.}
\institute{New York University (NYU), Electrical \& Computer Engineering Department, Brooklyn, NY 11201, USA. \and
NYU Abu Dhabi, Electrical Engineering, Abu Dhabi, UAE. \and
Center for Artificial Intelligence and Robotics, NYU Abu Dhabi, UAE. \and
Embodied AI and Robotics (AIR) Lab, NYU Abu Dhabi, UAE.
}
\maketitle
\begin{abstract}
We introduce an innovative approach to advancing semantic understanding in zero-shot object goal navigation (ZS-OGN), enhancing the autonomy of robots in unfamiliar environments. Traditional reliance on labeled data has been a limitation for robotic adaptability, which we address by employing a dual-component framework that integrates a GLIP Vision Language Model for initial detection and an InstructionBLIP model for validation. This combination not only refines object and environmental recognition but also fortifies the semantic interpretation, pivotal for navigational decision-making. Our method, rigorously tested in both simulated and real-world settings, exhibits marked improvements in navigation precision and reliability.
\end{abstract}

\section{Introduction}

\begin{figure}[t]
    \centering
    \includegraphics[width=0.7\linewidth]{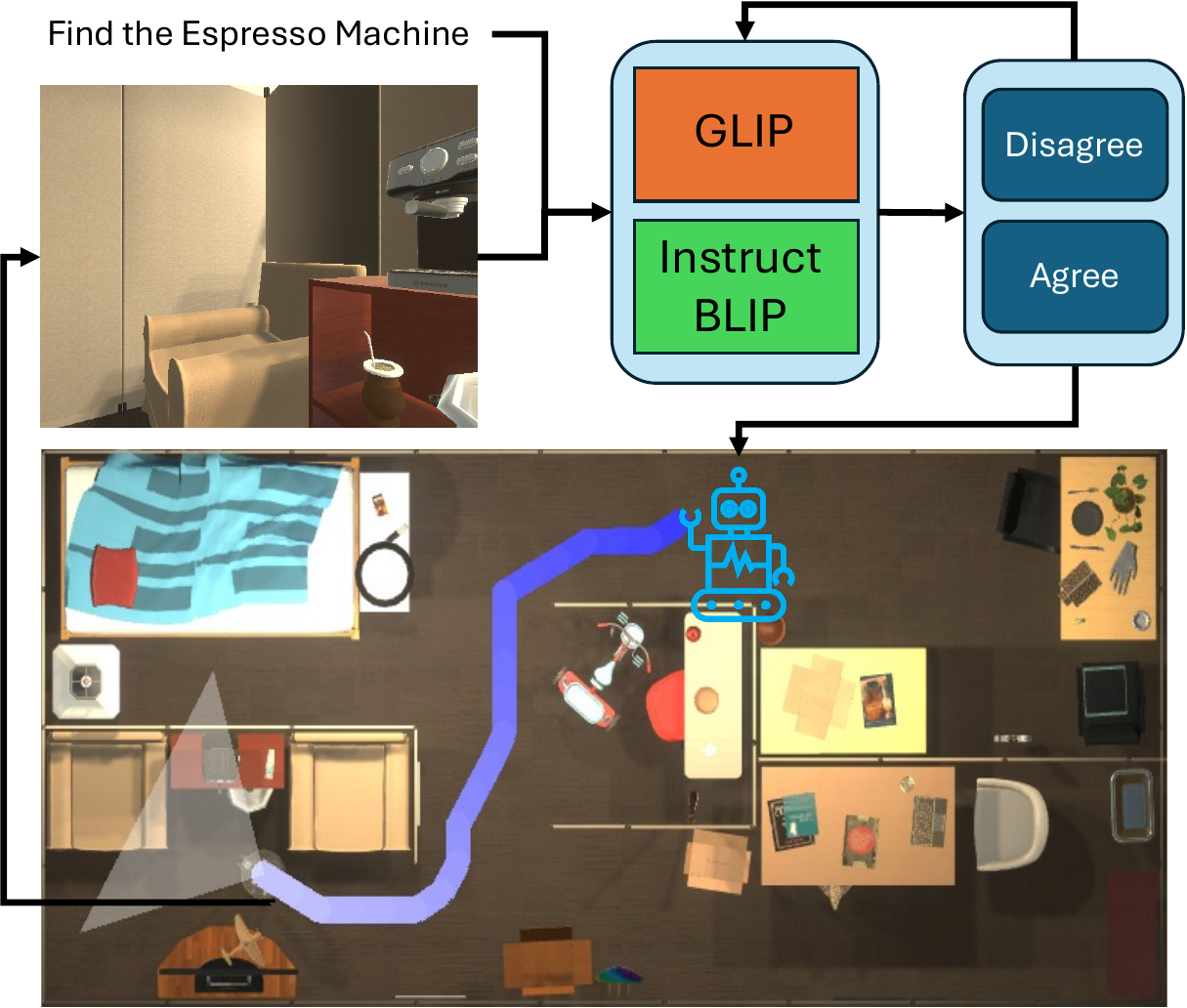}
    \caption{Illustration of the key component of our method ZS-OGN. The process begins with the GLIP Vision Language Model detecting the target object, in this case, an espresso machine. Subsequently, the InstructBLIP model evaluates the detection, either confirming the GLIP's proposal (`Agree') or not (`Disagree'), which influences the continuation or adjustment of the navigational plan.}
    \label{fig:intro}
\end{figure}

Object navigation is crucial for the autonomous operation of robots, which has traditionally depended on extensive labeled visual data. 
In Object Goal Navigation (OGN), the objective is to navigate uncharted environments in search of a specified, yet initially unseen, target object. Traditional methodologies in this domain have predominantly hinged on visual cues through either imitation \cite{silver2008high,karnan2022voila} or reinforcement learning \cite{kahn2018self,wohlke2021hierarchies} techniques, necessitating substantial data and annotations for effective training, thereby constraining their utility in diverse, real-world settings. 

This necessity has catalyzed a paradigm shift towards ZS-OGN strategies, designed to imbue robots with the capacity for immediate adaptation to novel objects and contexts \cite{zhao2023zero,majumdar2022zson,dorbala2023can,zhao2023semantic}. Zero-shot object goal navigation (ZS-OGN)\cite{chaplot2020learning,chaplot2020object,chen2022weakly,zheng2208jarvis,zhou2023esc,zhou2023navgpt,yao2022react} equips robots with the ability to identify and interact with objects they have not previously encountered, leveraging sensory inputs and exploratory behaviors \cite{majumdar2022zson}. This process allows robots to transcend the limitations of their training data, enhancing their versatility and expanding their potential for operation in dynamic and novel settings \cite{gadre2023cows,zhou2023esc}. Advancements in ZS-OGN bring us closer to developing robots that can comprehend and engage with a more diverse array of environments, representing a notable evolution in the field of robotics.

The existing ZS-OGN framework can be broken down into distinct components: semantic understanding, high-level exploration, and low-level navigation \cite{yamauchi1997frontier,gadre2023cows,zhou2023esc,shah2023navigation}. Semantic understanding is crucial as it provides the robot with the ability to discern unseen objects through environmental observations, and it lays the groundwork for subsequent exploration and navigation strategies. This underscores its critical role and highlights the significance of advancing semantic understanding within ZS-OGN.

Addressing the inherent limitations of direct visual embedding reliance, as observed in the ZER framework's \cite{al2022zero} two-stage process which begins with an ImageNav agent's foundational training, subsequent innovations have pivoted towards the exploitation of multimodal, semantic embedding spaces. This shift not only facilitates the comprehension of objects articulated in natural language but also circumvents the semantic void typical of image-goal embeddings bereft of semantic annotations. Alternative sensor modalities, such as LiDAR point cloud data, can enhance the semantic scene understanding via geometry-based reasoning \cite{komorowski2021minkloc3d,xia2024text2loc}.

To enhance the semantic knowledge of the robot within its environment, the integration of vision-language models into robotic systems has proved highly effective. These models merge the perceptual power of visual data with the deep contextual insights offered by linguistic information, creating a robust framework for interpreting complex environments. A notable innovation in this domain is VLMaps \cite{huang2023visual}, which integrates pre-trained visual-language features with 3D environmental reconstructions to improve spatial and semantic understanding and facilitate more natural interaction with the environment. The versatility of vision-language models in ZS-OGN \cite{gadre2023cows,majumdar2022zson} is further demonstrated by the use of CLIP \cite{gadre2022clip}, a pre-trained model that excels in diverse navigation scenarios. Moreover, ESC \cite{zhou2023esc} proposes a commonsense reasoning for an efficient frontier selection during robot exploration and enhances the CLIP-based model using a GLIP \cite{li2021grounded}, which represents a significant leap in scene comprehension. These advancements highlight the essential role of vision-language models in advancing the semantic understanding aspect of ZS-OGN and pushing robotic navigation toward higher levels of intelligence and adaptability. Despite such progress, the success rates for ZS-OGN approaches have not been fully satisfactory, often due to errors in object detection—either false identifications of visible goal objects or mistaken detections of nonexistent ones. This highlights the need for a more reliable semantic scene understanding framework that does not rely on further training. Addressing this gap, our research aims to develop a robust semantic understanding framework to reduce detection errors and enhance the precision and effectiveness of robot navigation.

Further advancements, as demonstrated by EmbCLIP and CoW \cite{khandelwal2022simple,gadre2023cows}, incorporate CLIP for enhanced vision-and-language navigation and object goal navigation, respectively, leveraging Frontier-Based Exploration (FBE) \cite{yamauchi1997frontier} to effectively demarcate and traverse the boundary between explored and unexplored territories. Such methodologies underscore a significant leap forward in navigating autonomously through open-world settings with unprecedented efficiency and adaptability.

Many of the aforementioned papers conduct studies in simulated or pre-recorded data. The common benchmarks and datasets provide photorealistic data, and the rate of improvement in the field is astounding. However, the real-life realizations of such systems are lacking. Attempts to run the proposed algorithms require many additional considerations for safety, one of which is the lack of accurate positioning information in indoor environments. To that end, we adapted parts of the ZS-OGN framework to be resilient to real-world uncertainty, noise, and latency, by integrating a safe global path planner and a local planner with dynamic obstacle avoidance. To the best of our knowledge, this work is the first in a ZS-OGN framework deployed in a real-life task.

In this paper, we present a novel approach to improve the semantic understanding of ZS-OGN through a two-part semantic pipeline, as shown in Figure \ref{fig:intro}, and demonstrate its efficacy in both simulated and real-world scenarios. The proposed framework consists of the GLIP Vision Language Model (VLM), responsible for the initial object and context detection, and InstructionBLIP, a validating VLM, to verify GLIP's detections. These components collaboratively enhance the accuracy of identifying objects and their surroundings. 

The contributions of this work are:
\begin{itemize}
    \item  A ``Doubly Right" semantic understanding framework for Zero-Shot Object Goal Navigation (ZS-OGN), which employs a dual verification system to enhance the reliability and accuracy of object detection in unfamiliar environments,
    \item  State-of-the-art performance on common simulation platforms, indicating that the proposed framework effectively improves the semantic understanding necessary for ZS-OGN,
    \item Implementation of a safety-optimal path planning algorithm to minimize chances of collision in real-world conditions, and 
    \item Experimental validation of the ZS-OGN pipeline in a real-world apartment
\end{itemize}

The rest of the paper is structured as follows: Problem details are provided in Section~\ref{sec:problem-statement}. Implementation details of the proposed system are given in Section~\ref{sec:approach}, and simulation and real-world study results are shown in Section~\ref{sec:simulation-studies} and \ref{sec:experimental-studies}, respectively.

\section{Problem Statement}
\label{sec:problem-statement}
In ZS-OGN tasks, a robot is tasked with locating the target object, denoted as \( g_i \), which it has not previously encountered within an unexplored environment \( s_i \), and this must be accomplished without prior navigational data training. At each timestep \( t \), the robot captures a color image \( I_t \), depth data \( d_t \), and its own pose—comprising its coordinates \( (x_t, y_t) \) and heading \( \theta_t \). The robot integrates its pose data over time to compute its current location. Utilizing the data from each timestep, the robot selects an action \( a \) from a set of possible actions \( \mathcal{A} \) which includes actions such as advancing, rotating left, rotating right, and halting. The navigation process is deemed successful when the robot elects to halt within a specified proximity to the target object.

\section{Approach}
\label{sec:approach}

\begin{figure}[t]
    \centering
    \includegraphics[width=0.9\textwidth]{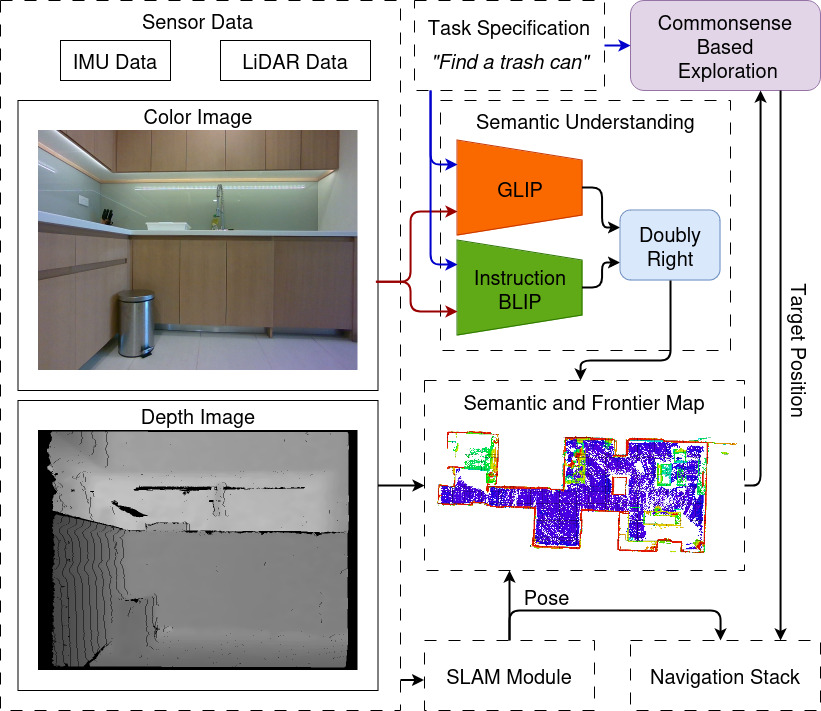}
    \caption{The flow of our proposed method in a real-world scenario.}
    \label{fig:pipe}
\end{figure}

In this section, we delineate our framework for ZS-OGN. 
As depicted in Figure~\ref{fig:pipe}, our approach incorporates a frontier-based exploration method, which is widely recognized in the field of ZS-OGN. The process initiates with the transformation of the input image into semantic data, subsequently integrating this information into a semantic map. The framework then harnesses the commonsense reasoning capabilities of large language models to determine the subsequent frontier for exploration.  Next, the Fast Marching Method is employed to compute the shortest path from the agent's current location to the designated target. Lastly, our innovative `Doubly Right' semantic understanding framework is applied to verify the accurate detection of the target object.

\subsection{Doubly Right Semantic Understanding}
In developing a robust framework for ZS-OGN, we introduce a novel method called "Doubly Right," which incorporates a dual verification system using Vision-Language Models (VLMs) to mitigate common detection errors. This approach, provided in Algorithm~\ref{alg:doubly-right}, seeks to increase the reliability of object detection and room recognition within navigational tasks.

The process begins with an Initiator VLM. Following the ESC framework, we employ the Grounded Language-Image Pre-training (GLIP) model for the preliminary identification of objects and rooms in environmental representations. The GLIP model processes visual inputs using zero-shot learning capabilities to generalize its detection beyond the training data through natural language prompts. When the GLIP model detects an object or room type, it assigns a provisional label and triggers the Validator VLM, InstructionBLIP, to assess the detection for accuracy.
For each visual input \( I_t \) at time \( t \), the framework executes an initial detection with the GLIP model:

\begin{equation}
O_{init} = \text{GLIP}(I_t, P_o),
\end{equation}
where \( O_{init} \) denotes the set of detected target objects by GLIP, and \( P_o \) represent the respective object prompting phrases applied to GLIP. Regardless of whether \( O_{init} \) is an empty set or contains detections, the Validator VLM, InstructionBLIP, then reviews these initial findings. It cross-references the output from the GLIP with the task instructions to confirm their validity. If InstructionBLIP identifies a need for reassessment, it advises that the environment be re-examined, indicating potential discrepancies in the initial detection. The algorithm will then return to the Initiator VLM to conduct a reevaluation. In cases where InstructionBLIP does not advise further inspection, a flag $Goal$ is set to \texttt{True}, indicating that the target object has been reliably detected and the navigational task is complete. This flag's status is critical as it confirms the end of the navigational sequence, ensuring that erroneous detections are addressed and the accuracy of the navigational decision-making is improved.

The validation process serves as a double-check mechanism, confirming the detected objects and their spatial contexts are indeed relevant and accurate for the navigational task at hand. By implementing this two-fold verification strategy, our framework aims to reduce detection errors that impede the success of autonomous navigation systems. The approach ensures that navigation decisions are based on a reliable semantic understanding of the environment, enhancing the system's performance in novel or previously unseen settings.

\begin{algorithm}
\caption{Doubly Right: Zero-Shot Object Goal Navigation Framework}
\label{alg:doubly-right}
\begin{algorithmic}[1]
\State \textbf{Input:} Visual input $I_t$ at time $t$, object prompting phrases $P_o$, validation prompting $P_v$
\State \textbf{Output:} a flag $Goal$ indicates whether the target object is found.
\State
\State \textbf{Initiator VLM:} Employ GLIP model following the ESC framework
\State $O_{init} \gets \Call{GLIP}{I_t, P_o}$ \Comment{Detect objects and rooms with GLIP}
    \State \textbf{Validator VLM:} Invoke InstructionBLIP model

    \If{$\Call{InstructionBLIP}{I_t,O_{init},P_v}$ advises reassessment}
        \State \textbf{Goto} Initiator VLM \Comment{Reassess the detections}
    \Else
        \State $Goal=\texttt{True}$
    \EndIf
\end{algorithmic}
\end{algorithm}

\subsection{Semantic and Frontier Map}

Our approach for ZS-OGN follows the ESC framework to construct a semantic navigation map critical for autonomous navigation. Utilizing depth input \(d_t\) and the agent’s 2D pose 
$\big[\begin{matrix}x_t & y_t & \theta_t\end{matrix}\big]^\top \in \mathbb{R}^2 \times \mathbb{S}$, we generate a foundational 2D navigation map. The GLIP model is then applied to enrich this map semantically by detecting objects and room types via zero-shot learning:

\begin{equation}
S_{map} = f(d_t, x_t,y_t,\theta_t, \mathcal{R}_i),
\end{equation}
where \(S_{map}\) represents the semantic map, \(d_t\) is the depth input at time \(t\), and \(\mathcal{R}_i\) symbolizes the environmental representation obtained from the GLIP model including room types and commonly occurring objects.

Then, the frontier map is constructed to delineate the boundaries for exploration. The process begins by identifying the periphery of the unoccupied area within the navigation map, forming candidate frontiers between free and unknown spaces \cite{ramakrishnan2022poni}.
These candidates are scrutinized by the Commonsense Policy $\pi_{cs}$ to prioritize the next exploration target:

\begin{equation}
F_{t} = \pi_{cs}\left( \widehat{F} \right),
\end{equation}
where \(\widehat{F}\) denotes the set of candidate frontiers, and \(F_{t}\) is the selected target frontier.

The semantic navigation map \(S_{map}\) is critical for the understanding of environmental semantics, enhancing the robot's interaction with the environment. Concurrently, the frontier map guides the exploratory progression, enabling systematic and informed exploration of new areas. Together, these maps form a comprehensive navigation system that supports autonomous agents in complex and dynamic settings.

\subsection{Commonsense Policy for Exploration}

Our navigation framework assimilates a commonsense reasoning module, adapted from the established ESC framework, to complement the autonomous navigation process. This module leverages the semantic understanding derived from our semantic navigation map to make contextual inferences, enhancing navigational decision-making.

The adopted ESC commonsense module $\text{ESC}_{cs}$ analyzes spatial relationships and object functionalities, drawing on a knowledge base of object-room correlations and navigational heuristics. Such inferences enable the anticipation of environmental elements indirectly indicated by the current sensory data:

\begin{equation}
C_{i} = \text{ESC}_{cs}(S_{map}),
\end{equation}
where \(C_{i}\) denotes the reasoned inferences based on the semantic map \(S_{map}\). These inferences feed into the Global Commonsense Policy, informing frontier selection for targeted exploration.

By incorporating the ESC commonsense reasoning, the framework attains a sophisticated level of environmental interpretation, pivotal for navigating through dynamic spaces. This integration, although not our core innovation, significantly enhances the overall efficacy of the navigation system.

\section{Simulation Studies}
\label{sec:simulation-studies}
\subsection{Dataset}
\paragraph{\textbf{RoboTHOR}}
The RoboTHOR dataset has been developed to validate navigation systems within authentic real-world scenarios. This benchmark features 89 meticulously designed apartment scenes, augmented by a comprehensive collection of 731 unique objects. Adhering to protocols from previous studies \cite{gadre2023cows,zhou2023esc}, our proposed method is subjected to an assessment comprising more than 1,800 validation episodes across 15 different environments. The evaluation is focused on 12 principal object categories crucial for navigation.

\paragraph{\textbf{PASTURE}}
The Pasture dataset augments RoboTHOR’s validation scenarios with additional object variations and complexity across 2,520 navigation tasks, enhancing the robustness of navigational model testing. It introduces an intricate mix of object sizes, colors, and materials, alongside detailed spatial relationships, to create a more challenging benchmark that closely mimics real-world conditions.

\subsection{Metrics}
Consistent with established benchmarks in the field, we utilize Success Rate (SR) and Success Weighted by Path Length (SPL) to evaluate agent performance. SR measures the agent’s accuracy in reaching the target within a meter's distance, presented as a percentage, where a higher rate indicates better performance. Conversely, SPL gauges the efficiency of the navigation, comparing the agent’s actual traveled path with the ideal shortest path, thus reflecting the agent's navigational efficacy and the optimization of its chosen route.

\subsection{Baselines}
In our experiment, we measure the performance of our method against baseline models such as CoW and ESC. CoW, designed for Zero-Shot Object Navigation (ZS-OGN), leverages CLIP for dynamic object detection, enabling localization without prior navigational training. We further evaluate CoW's efficacy by comparing it with its variants that utilize different CLIP-based localization strategies, namely CLIP-Ref, CLIP-Patch, CLIP-Grad, MDETR, and OWL. Additionally, we assess ESC, which incorporates commonsense knowledge into navigation actions through a pre-trained vision and language model, enhancing the agent's ability to navigate and reason about objects and rooms in unseen environments.

\subsection{Results}

\setlength{\tabcolsep}{3pt}

\begin{table}
\centering
\caption{Zero-shot object goal navigation results on PASTURE\cite{gadre2023cows} benchmarks.}

\label{tab:pasture}
\resizebox{\linewidth}{!}{
\begin{tabular}{lccccccccc}
\toprule
& 
\multirow{2}{*}{\textbf{Uncom.}} &
\multirow{2}{*}{\textbf{Appear.}} &
\multirow{2}{*}{\textbf{Space}} &
\multirow{2}{*}{\shortstack{\textbf{Appear.} \\ \textbf{distract}}} &
\multirow{2}{*}{\shortstack{\textbf{Space} \\ \textbf{distract}}} &
\multirow{2}{*}{\textbf{Hid.}} &
\multirow{2}{*}{\shortstack{\textbf{Hid.} \\ \textbf{distract}}} &
\multicolumn{2}{c}{\multirow{2}{*}{\textbf{Average}}} \\
& & & & & & & & & \\

\multicolumn{1}{c}{\textbf{Method}} & SR & SR & SR & SR & SR & SR & SR & SPL & SR \\

\midrule

CoW CLIP-Ref. & 3.6 & 2.8 & 2.8 & 3.1 & 3.3 & 4.7 & 5.0 & 1.7 & 2.5 \\
CLIP-Patch & 18.1 & 13.3 & 13.3 & 10.8 & 10.8 & 17.5 & \textbf{17.8} & 9.0 & 14.2 \\
CLIP-Grad. & 16.1 & 11.9 & 11.7 & 9.7 & 10.3 & 14.4 & 16.1 & 9.2 & 12.9 \\
MDETR & 3.1 & 7.2 & 5.0 & 7.2 & 4.7 & 8.1 & 8.9 & 5.4 & 6.3 \\
OWL & 32.8 & 26.9 & 19.4 & 19.4 & 16.1 & 19.2 & 15.8 & 12.6 & 21.1 \\
Ours & \textbf{33.0} & \textbf{29.0} & \textbf{22.4} & \textbf{19.7} & \textbf{18.2} & \textbf{20.8} & 17.6 & \textbf{13.7} & \textbf{23.0} \\

\bottomrule

\end{tabular}
\label{table:res}
}
\end{table}

The performance data in Table \ref{tab:pasture} reveals that our method leads with an average Success Rate (SR) of 23.0\% and Success Weighted by Path Length (SPL) of 13.7, indicating effective and efficient navigation in diverse scenarios. It surpasses others, particularly in challenging categories involving uncommon objects and hidden distractions, underscoring its robustness and sophisticated semantic understanding. Notably, the OWL model also demonstrates commendable SRs, especially in environments with spatial distractions. In contrast, the CoW model and other CLIP-based methods display more modest performance, highlighting the complexities these models face in the richly varied navigation tasks of the Pasture dataset, which is designed to evaluate navigation models with its intricate array of objects.
In addition to the evaluation on the Pasture Dataset, We conduct further experiments on the RoboTHOR dataset. The results on the RoboTHOR dataset, as presented in Table \ref{tab:robothor} reveal that our method stands out with superior performance, achieving the highest Success Weighted by Path Length (SPL) at 18.3 and Success Rate (SR) at 35.2, demonstrating exceptional navigation efficacy among the 12 crucial object categories across 15 environments. 

\begin{table}
\centering
\caption{Zero-shot object navigation results on RoboTHOR \cite{deitke2020robothor} benchmarks. * denotes the reproduced result using the official implementation.}
\begin{tabular}{lcc}
\toprule

& \multicolumn{2}{c}{\textbf{Performance}} \\
\multicolumn{1}{c}{\textbf{Method}} & SPL & SR \\

\midrule

CLIP-Ref.\cite{gadre2023cows} & 1.0 & 1.8 \\
CLIP-Patch.\cite{gadre2023cows} & 7.7 & 15.3 \\
CLIP-Grad.\cite{gadre2023cows} & 7.4 & 12.1 \\
MDETR.\cite{gadre2023cows}& 8.4 & 9.9 \\
CLIP-OWL. \cite{gadre2023cows}& 13.4 & 21.9 \\
ESC* \cite{zhou2023esc} & 18.2  & 34.5  \\
Ours & \textbf{18.3}&\textbf{35.2} \\

\bottomrule

\end{tabular}
\label{tab:robothor}
\end{table}

\section{Experimental Studies}
\label{sec:experimental-studies}

The proposed ZS-OGN pipeline was validated in real-world scenarios on real robot hardware. Following the procedures from the simulation environments and benchmarks, the robotic agent was tasked with navigation near various household objects with no prior information about the environment or the actual object.

Additional modifications were necessary to allow the system to operate in a real-world environment to improve overall system robustness and safety in the face of noise and uncertainties. All such modifications are mentioned in their respective sections below.

\subsection{Environment}

\begin{figure}[ht]
    \centering
    \includegraphics[keepaspectratio, width=0.8\linewidth]{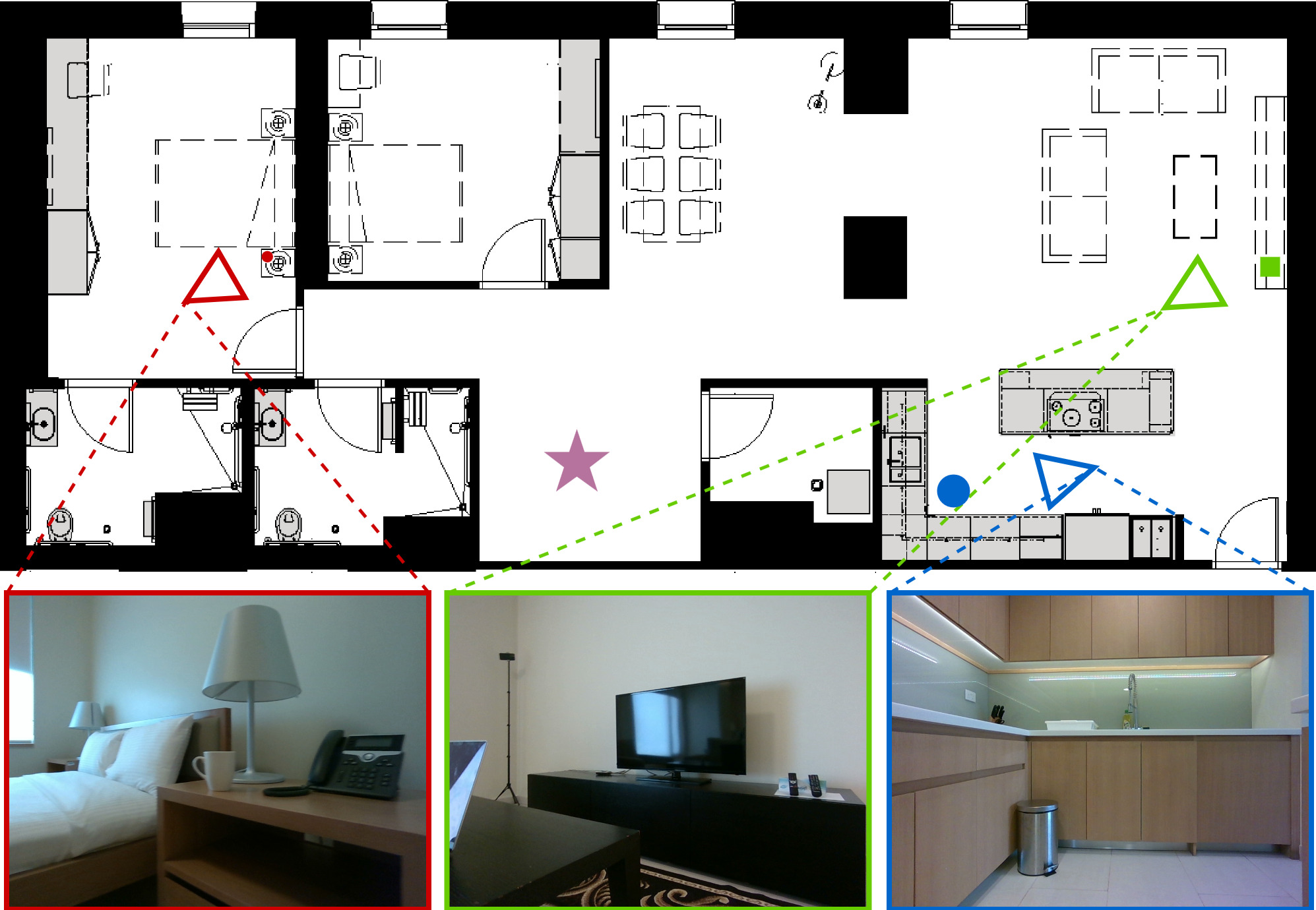}
    \caption{Layout of the apartment used in the experiment, along with the first-person views of the detected mug (red), remote (green), and trash can (blue). The starting location is marked with a star.}
    \label{fig:layout}
\end{figure}

The layout of the apartment in which the tests are performed is provided in Figure~\ref{fig:layout}, along with the locations of sample items to be detected and the first-person view from the robot when they are detected.

The placement of the objects was guided by common sense: a TV remote is expected to be found near the TV in a living room, and a garbage can can be found in a bathroom or a kitchen.

\subsection{Robotic Platform}

\begin{figure}
    \centering
    \includegraphics[keepaspectratio, width=0.5\linewidth]{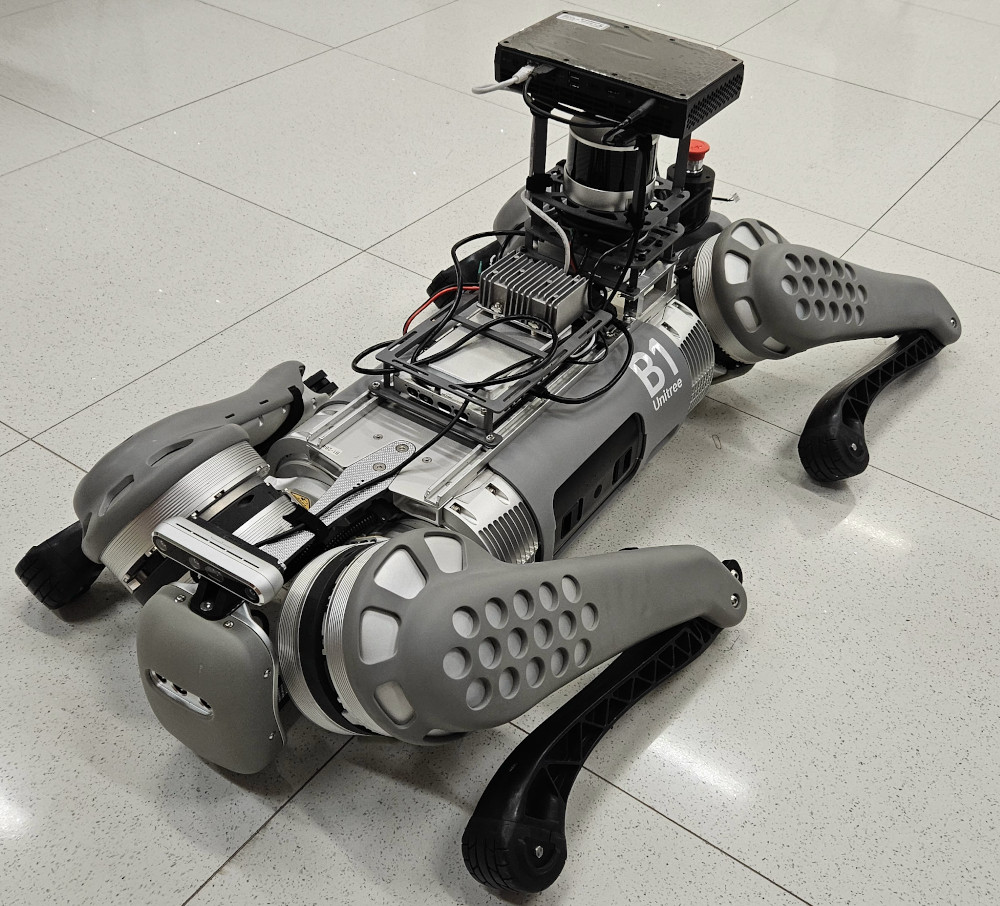}
    \caption{A photo of the Unitree B1 robotic platform.}
    \label{fig:robot}
\end{figure}

For the study, a Unitree B1 quadruped robot, equipped with a LiDAR (Pandar XT16) and an RGBD camera with IMU (Realsesne D455) was used. Simultaneous localization and mapping (SLAM), path planning, navigation, and low-level control were executed entirely on the robot's internal CPU, whereas the commonsense module for identifying goal location was run on the additional computer, attached to the robot. A photo of the vehicle is provided in Figure~\ref{fig:robot}.

The platform additionally streams depth images from 5 extra depth cameras, inertial measurements from an internal IMU, and a proprioceptive odometry estimate. However, none of the aforementioned data is used for this study.

\subsection{Odometry and Mapping}

The related literature for ZS-OGN utilizes RGBD-based mapping schemes that assume the agent pose is always available. However, without external infrastructure (e.g. GNSS for outdoors, motion capture for indoors, or dead-reckoning in both) the agent pose is neither readily available nor always accurate.

RTAB-Map~\cite{labbe2019rtab} was selected as the main driver for pose estimation, mapping, and localization. The robotic platform uses the onboard LiDAR and IMU sensors to estimate its odometry using RTAB-Map's ICP odometry module. With the addition of color images from the RGBD sensor, RTAB-Map's SLAM module is used for localization and mapping with global loop closure identification. Pose estimation and mapping are restricted to 3DoF ($xy$-coordinates and yaw angle $\theta$) since the environment is a single-story indoor location. 

\subsection{Safe Path Planning}

Many of the ZS-OGN frameworks output an action from the action space, such as ``move forward" or ``turn left" to carry out the navigation task. While such schemes function well in simulated environments, in which the actions can be carried out instantaneously and precisely, the real-world interaction needs to perform a trade-off between speed and accuracy. discrete action spaces prevent the agents from executing complex maneuvers (e.g. go through a narrow opening for a robot with non-circular footprint).
Finally, inherent noise in sensors and estimation for robotic platforms necessitate enhanced safety precautions to prevent inadvertent collisions with the environment. While a distance-optimal or effort-optimal path could take the robot to its desired location with high efficiency, the robot usually needs to navigate close to the obstacles in the environment. Latency in state estimation, control, and/or the actual movement can lead to collisions.

To that end, a medial-axis-based path planning algorithm, as proposed in~\cite{unlu2023uav}, was adopted. In this scheme, the agent is directed to follow a path that is maximally distant from any obstacles in the environment. Contrary to the original implementation, the system was adapted to use a 2D cost map for planning, to allow the paths to extend the unknown space, albeit at a high cost.

Let $\mathcal{M} \subseteq \mathbb{Z}_+^2$ denote the coordinates of the 2D occupancy grid map, with the occupancy encoded as an integer through a mapping $C: \mathcal{M} \rightarrow [0, 255]$. Since the cost map obtained from RTAB-Map is metric, there exists a surjective mapping between the real, metric coordinates $p \in \mathbb{R}^2$ and the map coordinates $\bar{p} \in \mathbb{Z}_+^2$. Let $p_r$ ($\bar{p}_r$) and $p_g$ ($\bar{p}_g$) denote the metric (map) coordinate of the robot's current position and target location, respectively.

The traversable set is a subset of the 2D cost map that does not map to a lethal cost (i.e. fully occupied or dangerous to approach) in the original cost map. The exact encoding may differ between different implementations, but defining the lethal cost set as $C_l \subset [0, 255]$, the traversable region is defined formally as
\begin{equation}
\mathcal{M}_t = \{ p \in \mathbb{Z}_+^2,\ C(p) \notin C_l \},
\end{equation}
with a cost to traverse the coordinate determined via the mapping $C$. The traversable set need not be connected. However, only the portion of the traversable set that contains the current coordinate of the robot is of interest, which will be denoted as $\mathcal{M}_t$ for simplicity.

The medial axis for this context is defined as the set of coordinates $\mathcal{S} \subseteq \mathcal{M}_t$ that are equidistant to multiple closest points in the boundary of the traversable region. Many implementations exist, but the thinning algorithm implementation in OpenCV~\cite{zhang1984fast} is used over a binary image representation in this work.

Finally, let $\bar{p}_\text{entry} \in \mathcal{S}$ be the closest coordinate to $\bar{p}_r$ with a linear path completely contained in $\mathcal{M}_t$. Similarly, let $\bar{p}_\text{exit} \in \mathcal{S}$ be the closest coordinate to $\bar{p}_g$ in the same fashion.

Given the above definitions, the path planning algorithm generates a path for navigation from the current position to the target position in 3 different segments:
\begin{itemize}
    \item A linear path $\mathcal{P}_\text{entry}$ from $\bar{p}_r$ to $\bar{p}_\text{entry}$,
    \item An ordered set $\mathcal{P}_\text{axis} \subseteq \mathcal{S}$, comprised of a series of adjacent coordinates that connect $\bar{p}_\text{entry}$ to $\bar{p}_\text{exit}$, and
    \item A linear $\mathcal{P}_\text{exit}$ from $\bar{p}_\text{exit}$ to $\bar{p}_g$.
\end{itemize}

\begin{figure}[ht]
    \centering
    \includegraphics[keepaspectratio, width=0.6\linewidth]{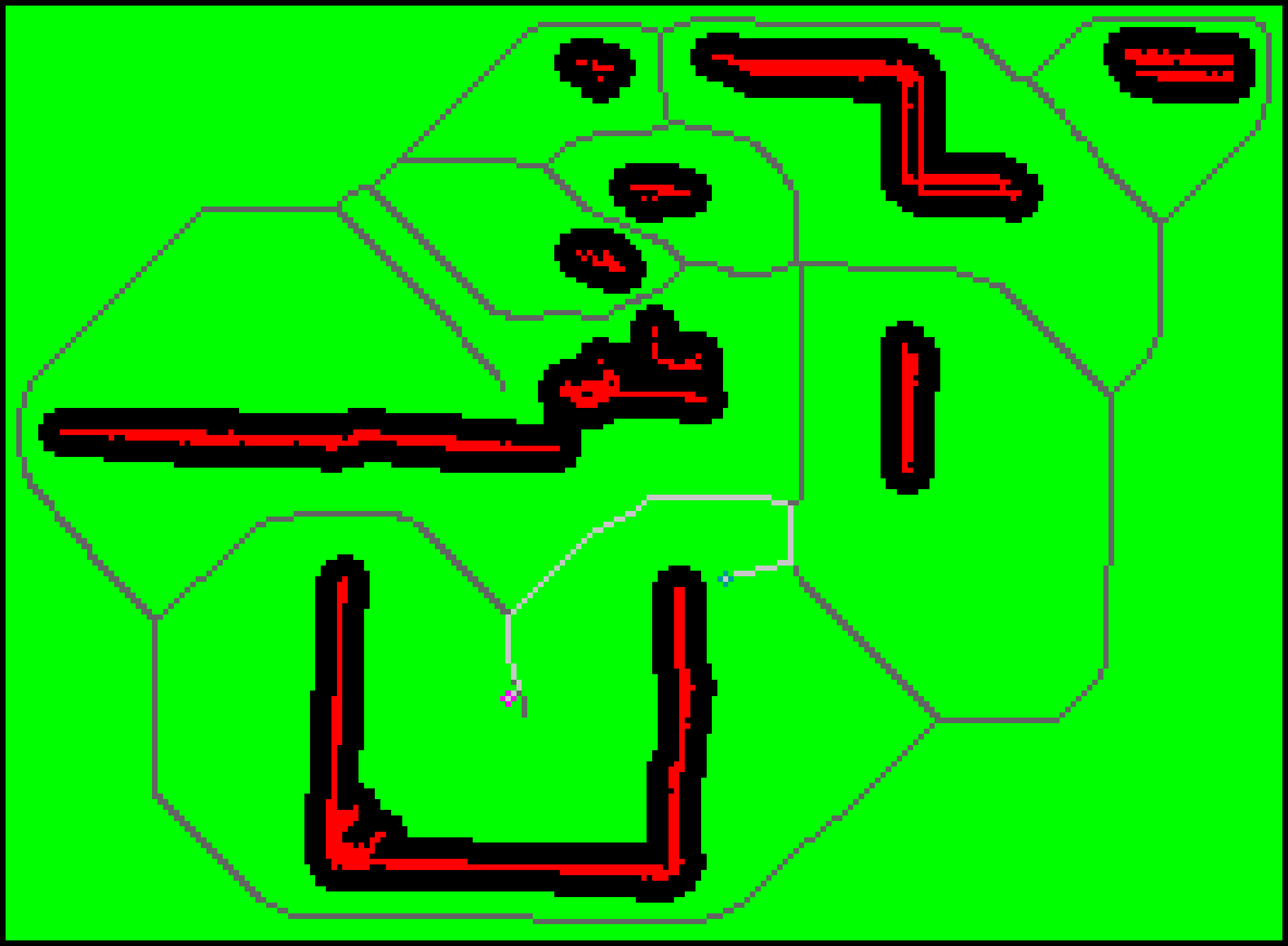}
    \caption{Sample path generated from the medial axis algorithm. The path (light gray) connects the current (magenta) and target (dark blue) coordinates through the medial axis (dark gray), entirely within the traversable region (green) and maximizing distance from obstacles (red). Black denotes lethal cost.}
    \label{fig:planner}
\end{figure}

A sample demonstration of the algorithm on a cost map computed during the experiment is provided in Figure~\ref{fig:planner}

\subsection{Implementation Details}

The path planning algorithm as proposed was implemented as a planner plugin within Nav2~\cite{macenski2020marathon2} framework, an open-source navigation stack with production deployments. 

Since the robot doesn't have a circular footprint, the cost map calculation, and in return the safe path planning, becomes challenging, as the orientation of the robot has an impact on its footprint. To avoid collisions and to follow the generated path, the DWB controller, a critic-based local planner with a dynamic window approach, is used. With basic tuning of DWB parameters and critics, the local and global planner combination was found to provide sufficient agility while preventing collisions.

\subsection{Results}

\begin{figure}[ht]
    \centering
    \includegraphics[keepaspectratio, width=0.7\linewidth]{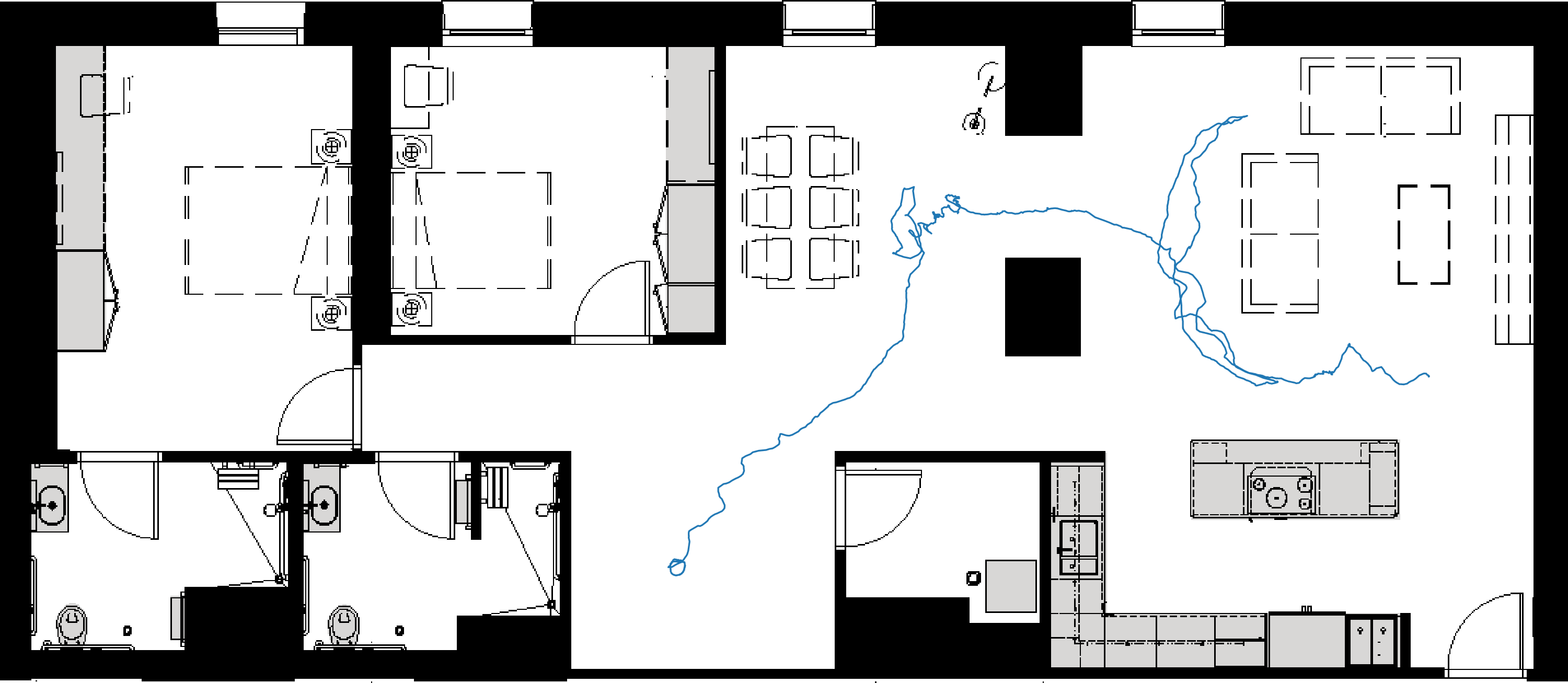}
    \caption{Robot path in searching for a TV remote, overlaid on the architectural plan of the apartment.}
    \label{fig:remote-overlay}
\end{figure}

\begin{figure}[ht]
    \centering
    \includegraphics[keepaspectratio, width=0.7\linewidth]{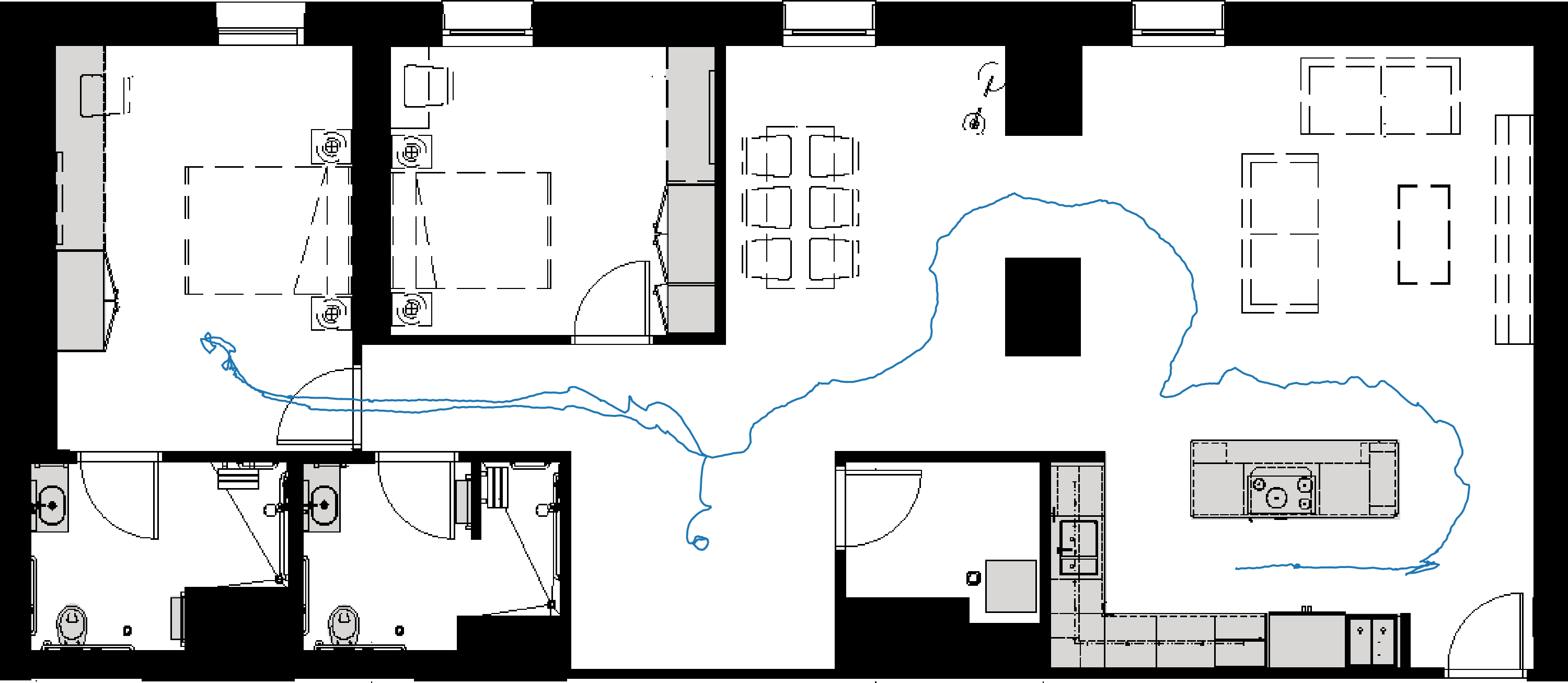}
    \caption{Robot path in searching for a trashcan, overlaid on the architectural plan of the apartment.}
    \label{fig:trashcan-overlay}
\end{figure}

The path taken by the robot to find a remote controller overlaid on the architectural plan of the apartment, is given in Figure~\ref{fig:remote-overlay}, and the path for finding a trashcan is provided in Figure~\ref{fig:trashcan-overlay}.

In finding the TV remote, the system appears to recognize the living room area, prioritizing exploration of the right-hand side of the apartment. Upon getting closer to the television, the remote controller was successfully recognized, and the mission was over.

For the setup to find a trash can, the robot initially attempts to navigate towards the bedroom on the left. At the time of exploration, the algorithm confused the featureless environment to be a bathroom and focused on its exploration. Eventually, the robot sees the bed and proceeds to focus on another frontier point. The robot observed the trashcan located in the kitchen initially, but the path planning algorithm proposed a more indirect approaching angle for the trashcan, in an attempt to minimize the cost of traversal. Eventually, the robot reached the trashcan, completing the task.

\section{Conclusion}
Our work presents the `Doubly Right' framework, a step forward in Zero-Shot Object Goal Navigation (ZS-OGN), enabling reliable semantic understanding in robotics. Our approach stands out as the first to implement ZS-OGN in real-world settings, demonstrating strong potential through simulation and practical application. Real-world tests demonstrated that lack of discerning features in the environment can result in the system making a poor initial choice in hindsight, due to the lack of knowledge about the observed environment, but the system is nevertheless able to correct course and complete the task. This breakthrough lays the groundwork for future autonomous systems to navigate novel environments without prior training, offering a glimpse into the next frontier of robotic adaptability and intelligence.

\bibliographystyle{splncs04}
\bibliography{references}
\end{document}